%% file: Main.tex
\documentclass[10pt,twocolumn,letterpaper]{article}

\usepackage{cvpr}              

\usepackage{graphicx}
\usepackage{amsmath}
\usepackage{amssymb}
\usepackage{booktabs}

\usepackage[pagebackref,breaklinks,colorlinks]{hyperref}

\usepackage[capitalize]{cleveref}
\crefname{section}{Sec.}{Secs.}
\Crefname{section}{Section}{Sections}
\Crefname{table}{Table}{Tables}
\crefname{table}{Tab.}{Tabs.}

\def\cvprPaperID{2649} 
\def\confName{CVPR}
\def\confYear{2022}

\begin{document}

\title{\emph{DArch}: Dental Arch Prior-assisted 3D Tooth Instance Segmentation \\
with Weak Annotations}

\author{
Liangdong Qiu\textsuperscript{\rm 1,3}\thanks{L. Qiu and C. Ye contribute equally.} 
\quad Chongjie Ye\textsuperscript{\rm 1}\footnotemark[1]
\quad Pei Chen\textsuperscript{\rm 1} 
\quad Yunbi Liu\textsuperscript{\rm 1,3} 
\quad Xiaoguang Han\textsuperscript{\rm 1,2}\thanks{Corresponding author}
\quad Shuguang Cui\textsuperscript{\rm1,2,3}\\
\textsuperscript{\rm 1}SSE, CUHK-Shenzhen
\quad \textsuperscript{\rm 2}FNii, CUHK-Shenzhen
\quad \textsuperscript{\rm 3}Shenzhen Research Institute of Big Data\\
{\tt\small \{liangdongqiu, chongjieye, peichen\}@link.cuhk.edu.cn, ybliu1994@gmail.com,}\\
{\tt\small\{hanxiaoguang, shuguangcui\}@cuhk.edu.cn}
}

\maketitle

\begin{abstract}
Automatic tooth instance segmentation on 3D dental models is a fundamental task for computer-aided orthodontic treatments. Existing learning-based methods rely heavily on expensive point-wise annotations. 
To alleviate this problem, we are the first to explore a low-cost annotation way for 3D tooth instance segmentation, \emph{i.e.}, labeling all tooth centroids and only a few teeth for each dental model. Regarding the challenge when only weak annotation is provided, we present a dental arch prior-assisted 3D tooth segmentation method, namely DArch. Our DArch consists of two stages, including tooth centroid detection and tooth instance segmentation. Accurately detecting the tooth centroids can help locate the individual tooth, thus benefiting the segmentation. Thus, our DArch proposes to leverage the dental arch prior to assist the detection. Specifically, we firstly propose a coarse-to-fine method to estimate the dental arch, in which the dental arch is initially generated by Bezier curve regression, and then a graph-based convolutional network (GCN) is trained to refine it. With the estimated dental arch, we then propose a novel Arch-aware Point Sampling (APS) method to assist the tooth centroid proposal generation. Meantime, a segmentor is independently trained using a patch-based training strategy, aiming to segment a tooth instance from a 3D patch centered at the tooth centroid. Experimental results on $4,773$ dental models have shown our DArch can accurately segment each tooth of a dental model, and its performance is superior to the state-of-the-art methods.
\end{abstract}


\input{section/intro}

\input{section/relatedwork}

\input{section/method}

\input{section/experiment}

\input{section/conclusion}

{\small
\bibliographystyle{ieee_fullname}
\bibliography{egbib}
}

\end{document}

%% file: section/intro.tex
\section{Introduction}
Dental models, obtained by direct intraoral scanning of the dentition, are commonly used in computer-aided dentistry. Computer-aided dentistry requires dental models as input to assist dentists to analyze and evaluate patient-specific dental health and dental arrangement for the following treatments. Automatic tooth instance segmentation on dental models is an essential prerequisite step for computer-aided orthodontic treatments. 

\begin{figure}[t]
  \centering
   \includegraphics[width=0.95\linewidth]{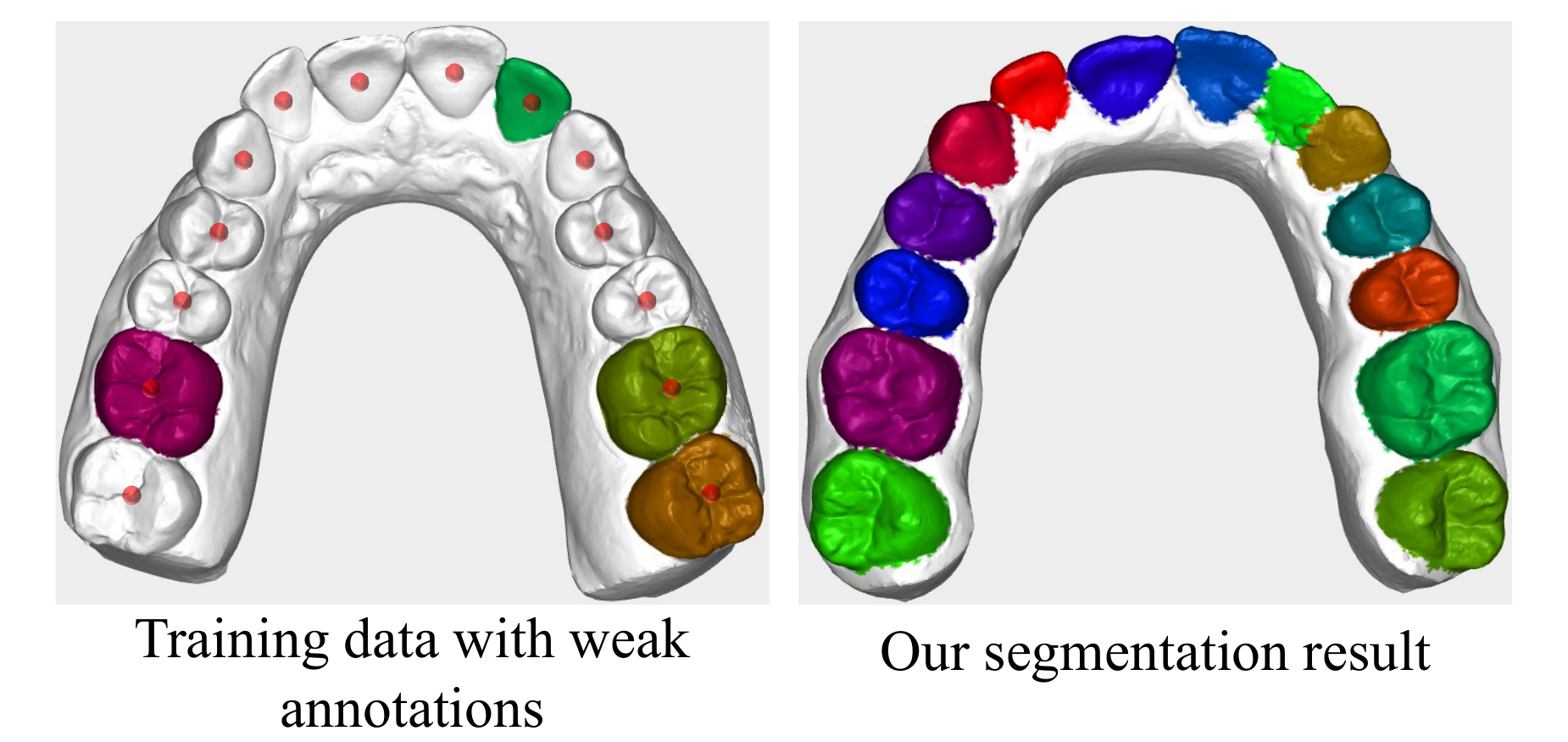}
   \caption{An example of our segmentation result when using training data with weak annotations for training models. Left: dental model with weak annotations, \emph{i.e.}, labelling all tooth centroids and only a few teeth. Right: our segmentation result.}
   \label{fig:teaser}
\vspace{-5mm}
\end{figure}

Although recent learning-based methods have achieved impressive performance on 3D tooth instance segmentation~\cite{zhang2020automatic, sun2020automatic, sun2020tooth}, they rely heavily on a large number of data with dense manual annotations, such as labeling all points of every individual tooth from a dental model.
Since annotating such training data is particularly time-consuming, it is hard to collect a large enough dataset to cover complex dental models in real-world, thus largely limiting the generalization of those learning-based segmentation methods~\cite{tian2019automatic, zanjani2019mask, xu20183d}. One of the main challenges in automatic 3D tooth instance segmentation is locating each tooth object on a variety of dental models, some of which have missing, crowding, or misaligned teeth. Cui et al.~\cite{cui2021tsegnet} found that in the tooth detection stage, the tooth centroid is a more reliable signal than the bounding box that is used to crop the detected tooth objects in the traditional approaches. Tooth detection thus can be converted to tooth centroids detection. Motivated by their work, we propose a feasible and low-cost annotation way as shown in the right of the Fig.~\ref{fig:teaser}, that is, \textbf{ specifying 3D centroids for all tooth instances and labeling dense instance mask for only a few teeth for each dental model}, to alleviate the demand for expensive point-wise annotations. 
In this paper, we present a novel \emph{detect-and-segment} framework, including detecting tooth centroids and segmenting every teeth instance assigned to the corresponding teeth centroid. We mainly focus on the detection stage based on the intuition that the more accurate the detection, the better the segmentation. Furthermore, we adopt a patch-based training strategy to decrease the discriminating difficulty for our segmentor, which aims to segment a tooth instance from a 3D patch centered at the tooth centroid, rather than segment all tooth instances from a whole point cloud data of a dental model.
In such a way, it makes a great demand on the tooth centroid prediction in our method. Previous detection methods for 3D point clouds~\cite{nezhadarya2019boxnet, cui2021tsegnet,qi2019deep} generally use the furthest point sampling (FPS) method to uniformly select the sampling points for generating proposals. For tooth centroid detection, the sampled points by FPS method generally contain irrelevant points, such as one located on the tooth crown and gingiva, which may lead to inaccurate proposals for tooth centroids. 

To accurately and completely predict each tooth centroid of a dental model, we propose an arch-aware point sampling (APS) module for tooth centroid detection by introducing dental arch prior to assist the detection procedure. This is based on the observations that a dental arch naturally depicts one's overall dentition, and all tooth centroids will fall on it. To estimate the dental arch of each dental model, we first formulate the dental arch by representing it as a curve passing through teeth centroids and then adopt a lightweight 1-D convolutional network to refine the dental arch~\cite{ling2019fast, peng2020deep}.
Different from FPS method that performs uniform sampling from the whole tooth votes, we sample points along the estimated dental arch to filter out a majority of irrelevant points. Experiments have shown that our proposed APS strategy can largely improve the detection accuracy for tooth centroids compared to FPS and benefit the following segmentation. 

To the best of our knowledge, this is the first attempt for 3D tooth instance segmentation on dental models with weak annotations. The main contributions of our work can be summarized as follows:

\begin{itemize}
    \item We are the first to explore a low-cost annotation way for 3D tooth instance segmentation and propose a novel framework named DArch to handle this challenging task with weak annotations. We hope this attempt will inspire more learning-based methods in the weakly-annotated scenario.
    \item We propose a coarse-to-fine method to estimate the dental arch. Specifically, the dental arch is initially approximated by Bézier curve regression, and then a graph-based convolutional network (GCN) is used for further refinement. 
    \item We introduce a dental arch-aware point sampling (APS) module for tooth centroid detection by introducing dental arch prior to assist the proposal generation. 
    \item Extensive experiments have shown that our proposed DArch can vastly improve the performance of tooth centroid detection compared to other methods using other sampling strategies. As for the segmentation performance, our DArch is superior to the state-of-the-art methods in both weakly- and fully-annotated scenarios.
\end{itemize}

%% file: section/relatedwork.tex
\begin{figure*}[ht]
\begin{center}
  \includegraphics[width=0.95\linewidth]{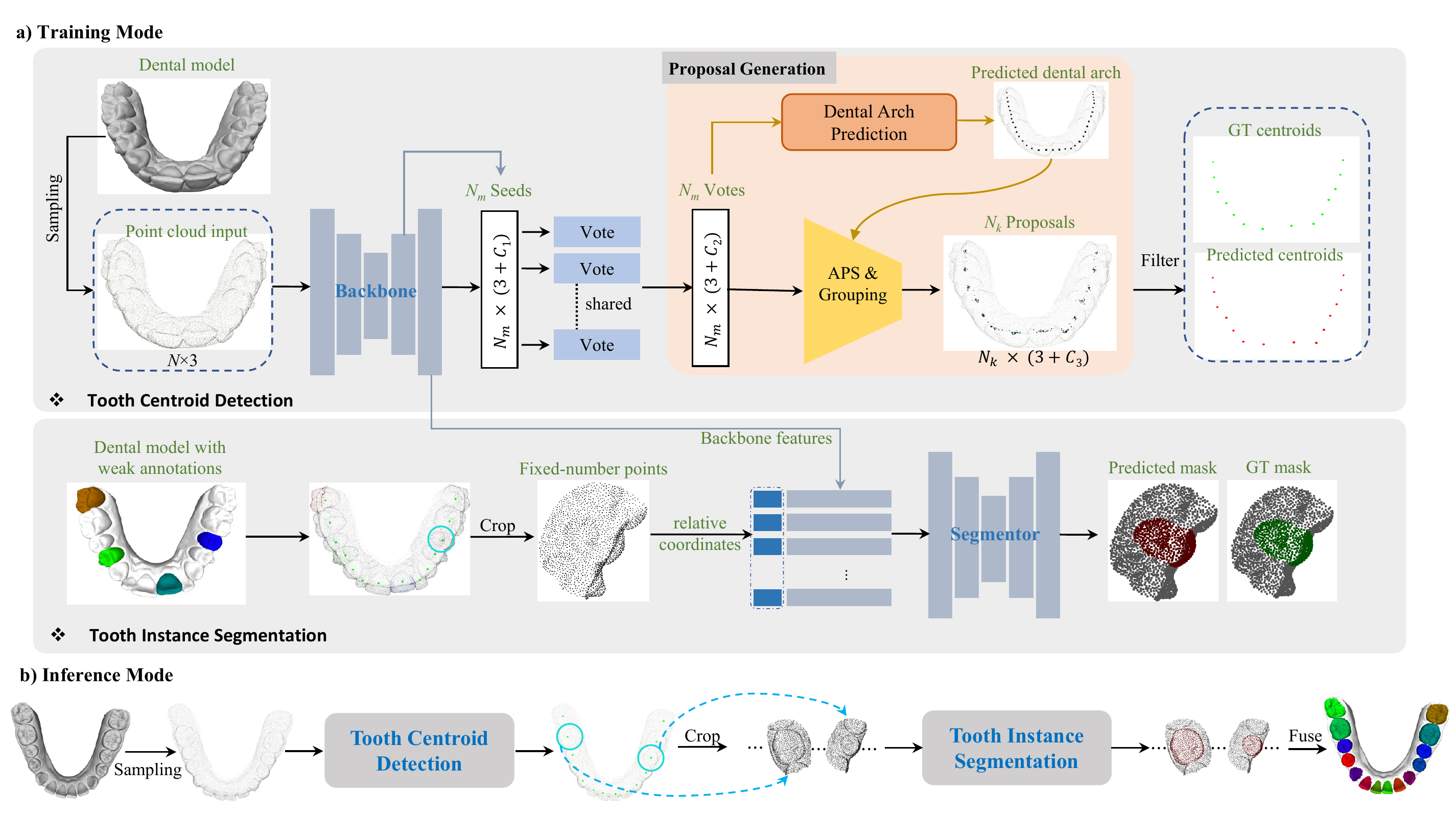}
\end{center}
  \caption{\textbf{Illustration of our DArch in both training and inference mode.} Our DArch consists of two parts of tooth centroid detection and tooth instance segmentation. In the inference mode, our DArch can segment all tooth instances by fusing the patch-based results. APS: Arch-aware point sampling.}
\label{fig:pipeline}
\vspace{-5mm}
\end{figure*}

\section{Related Work}
\subsection{3D Understanding in Natural Scene} 
3D understanding in natural scenes usually involves object detection~\cite{liang2018deep, yang2018pixor, song2016deep, pang20163d}, instance segmentation~\cite{han2020occuseg, jiang2020pointgroup}, shape understanding~\cite{wang2017cnn, wu20153d}, part segmentation~\cite{le2017multi, wang2019voxsegnet} and so on, which is a fundamental problem in computer vision. In recent years, some deep learning-based methods have been proposed on different representations, such as volumetric data~\cite{zhou2018voxelnet, zhou2018voxelnet, maturana2015voxnet, qi2016volumetric}, point cloud~\cite{xu2021paconv, shi2019pointrcnn, shi2020pv} and other representations~\cite{park2019deepsdf, riegler2017octnet}. A point cloud is among one of the most popular ways to represent the 3D shape or object. PointNet~\cite{qi2017pointnet} is an early representative attempt to design a novel deep network suitable for unordered point sets in 3D. 
PointNet++~\cite{qi2017pointnet++} and PointCNN~\cite{li2018pointcnn} extended PointNet by recursively applying it in a hierarchical fashion, so as to learn deep point set features efficiently and robustly. These two works inspired a lot of follow-up works. For example, VoteNet~\cite{qi2019deep} propose to detect 3D objects by endowing point cloud deep networks (\emph{i.e.,  PointNet++}) with a voting mechanism similar to the classical Hough voting. By voting, VoteNet essentially generates new points that lie close to objects centers, which can be grouped and aggregated to generate box proposals. Regarding the strong ability of feature representation of PointNet++ and the voting mechanism to generate objects centers in VoteNet, we adopt VoteNet as the basic architecture of our tooth centroid detection network and PointNet++ as the backbone network to exact the deep point features of the fine-grained tooth objects. To generate proposals from the votes in the proposal step, VoteNet used the furthest point sampling (FPS) to uniformly sample K vote clusters. Such a sampling strategy may select irrelevant vote clusters for tooth centroid detection, such as one located on the tooth crown and gingiva. To avoid this problem, instead of using FPS, we propose an arch-aware point sampling (APS) strategy to assist in generating proposals of tooth centroid by leveraging the dental arch prior.

 \subsection{3D Tooth Understanding} 
Recently, deep learning-based methods have been popularly used to handle the task of tooth instance segmentation~\cite{ tian2019automatic, zhang2020automatic, zanjani2019mask}. For example, Mask MCNet~\cite{zanjani2019mask} proposed a framework that combines the Monte Carlo Convolutional Network (MCCNet) with Mask R-CNN to simultaneously locate each tooth object by predicting its bounding box and segment all the tooth points inside the box. Graph convolutional neural network-based frameworks (GCN) ~\cite{sun2020automatic, zhang2021tsgcnet, sun2020tooth} have been proposed to learn more discriminative geometric features for 3D dental model segmentation. TSegNet~\cite{cui2021tsegnet} found that the tooth centroid is a more reliable signal than the bounding box in the tooth detection stage, and based on this observation proposed a novel pipeline that formulates the dental model segmentation as two sub-problems: robust tooth centroids prediction and accurate individual tooth segmentation on point cloud data. However, existing learning-based methods heavily depended on expensive dense point-wise annotations, that is, labeling all teeth of each dental model in the training data, to supervise the training process. Such a full annotation way brings a considerable burden for human labeling and increases the difficulty of collecting a large number of data, thus limiting these methods to real-world applications. In this paper, we are the first to study a 3D tooth instance segmentation problem with limited annotations. Motivated by those methods above, our proposed DArch includes a tooth centroid detection model to identify each tooth object and a tooth instance segmentation model
to segment every tooth instance. To accurately detect each tooth centroid, we propose to estimate the dental arch and leverage the estimated dental arch to assist the proposal generation of tooth centroids.

%% file: section/method.tex
\section{Method}
\subsection{Overview}
In this work, we propose a novel detect-and-segment framework, dubbed DArch, to tackle the challenging task of 3D tooth instance segmentation with weak annotations. Our DArch aims to segment all tooth instances given a point cloud input of a single dental model. As shown in Fig.~\ref{fig:pipeline}, our DArch consists of two parts, including tooth centroid detection and tooth instance segmentation. In particular, to accurately predict all tooth centroids, we introduce a dental arch prediction module to estimate the dental arch and propose an arch-aware point sampling (APS) strategy to generate the centroid proposals. Our segmentation network adopts a patch-based training strategy, and in the inference phase the trained segmentor can predict all the tooth instances from a dental model by fusing all patch-based segmentation results. We will elaborate our detection and segmentation networks as follows.
\subsection{Tooth Centroid Detection}
Our tooth centroid detection network consists of a detection backbone and a branch of arch generation. We adopt the VoteNet~\cite{qi2019deep} as our detection backbone regarding its solid architecture and voting mechanism. As we know, proposal generation is one of the most vital parts for 3D object center detection. Considering the fine-grained structures of teeth, we propose an APS method to replace the FPS sampling method used in VoteNet for generating tooth centroid proposals.
In the following, we first review the work of VoteNet briefly and then propose our arch generation method. Finally, we present our APS method for proposal generation.
\begin{figure}[t]
  \centering
   \includegraphics[width=1\linewidth]{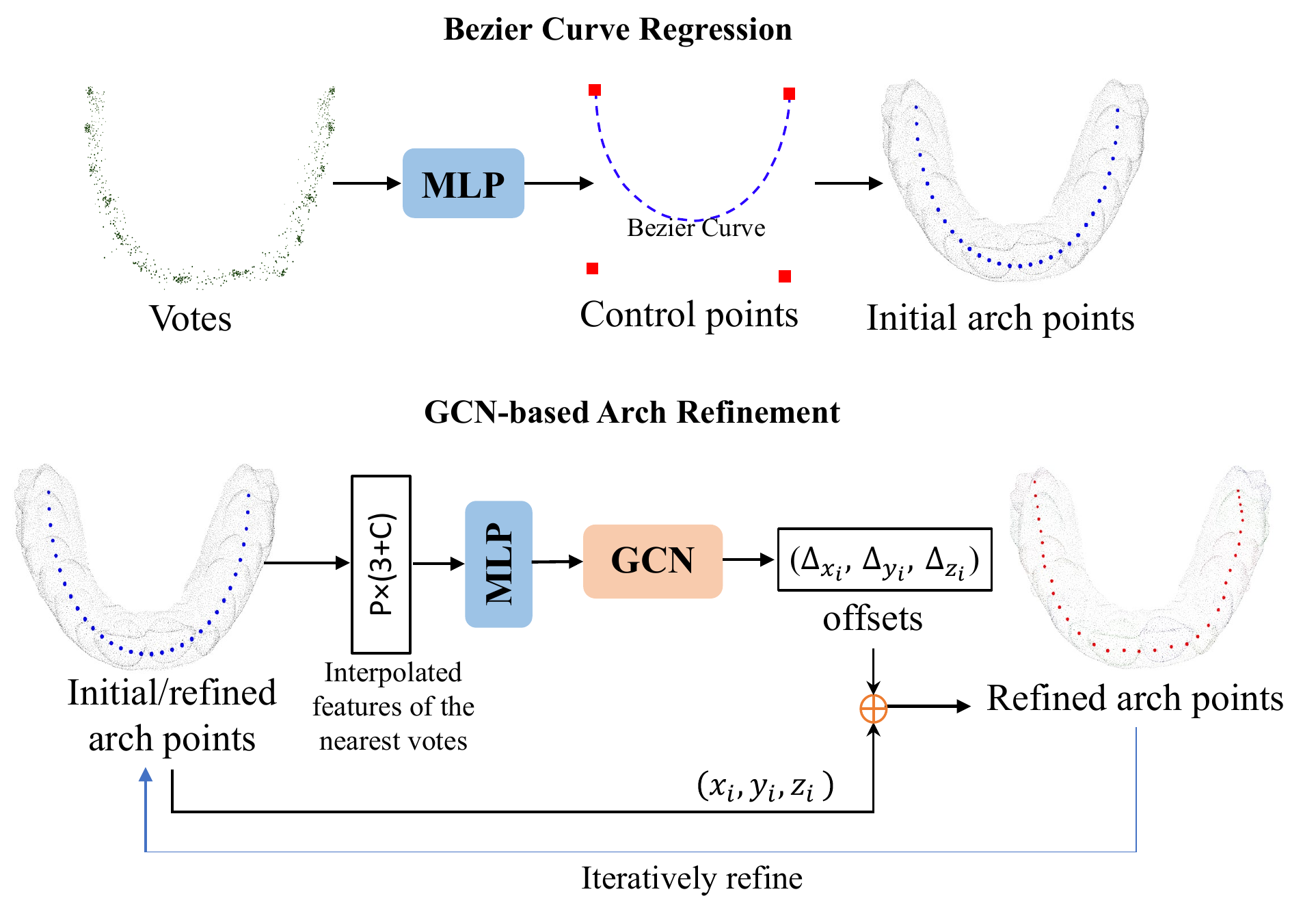}
   \caption{The overview of our proposed dental arch prediction method. Our method consists of two steps, including Bézier regression and GCN refinement. An initial curve is sampled from generated Bézier curve. Then, the points $\hat{\mathbf{X}}$ are refined by offsets iteratively.}
   \label{fig:arch}
\vspace{-5mm}
\end{figure}
\subsubsection{Review of VoteNet}
The original VoteNet is proposed by Qi et al.~\cite{qi2019deep}. It is tailored for 3D point cloud detection based on PointNet++~\cite{qi2017pointnet++}. Given a set of input 3D points $\left\{p_i\right\}_{i=1}^N$, the backbone network of PointNet++ selects seed points and generate enriched $C$-dimensional feature vector. The point coordinates are embedded into the $C$-dimensional feature vector to represent the seeds $\left\{s_i\right\}_{i=1}^M$, where $s_i$ is a $(3+C)$-dimentional feature vector. Then the seed points are fed into a shared multi-layer perception (MLP) to compute votes $\left\{v_i\right\}_{i=1}^M$. The generated voting $v_i$ will be aggregated around object center.

After generating votes $v_i$, FPS is used to sample a subset of the votes to get $\left\{v_i\right\}_{i=1}^K$. Then by finding all near votes within a certain Euclidean distance, votes are generated into $K$ clusters, followed by a three-layer MLP to generate the proposals. Finally, NMS is applied to filter out the overlapped proposals and generate the final prediction. 
The sampling method is very significant for generating reasonable proposals. For the specific task of tooth centroid detection, FPS may sample irrelevant points from the whole tooth votes, such as one located on the tooth crown and gingiva, due to its uniform and sparse sampling mechanism, resulting in inaccurate proposals. To address this issue, we propose to predict the dental arch that passes through all tooth centroids and then propose an APS method based on the predicted dental arch to replace with FPS for generating accurate proposals.

\subsubsection{Dental Arch Prediction}
The dental arch can describe the teeth arrangement of a dental model. 
To automatically predict the dental arch for each dental model, we propose a coarse-to-fine dental arch prediction method. 
As shown in Fig~\ref{fig:arch}, our proposed dental arch prediction method first roughly predicts the dental arch by regressing a cubic Bézier curve and then adopts a GCN-based network to refine the arch. In the following, we present our dental arch prediction method in detail.
\paragraph{Bézier Curve Regression}
Recently, it has been shown that the human dental arch form is accurately represented mathematically by the beta function~\cite{noroozi2001dental}. Motivated by ~\cite{noroozi2001dental}, we select a simple function, cubic Bézier curve, from the beta function set to initially approximate dental arch. The specific cubic Bézier curve can be decided by four control points. 
The ground truth of control points are obtained by minimizing the distance between the synthesized Bézier curve and the teeth centroids. As shown in the top of Fig.~\ref{fig:arch}, we use an MLP to predict $4$ control points
$\left\{x_i^{ctr}\right\}_{i=1}^4$.
The loss is defined as
\begin{equation}
  L_{ctr}=\frac{1}{4} \sum_{i=1}^{4} \ell_{1}\left(\hat{\mathbf{x}}_{i}^{ctr}-\mathbf{x}_{i}^{ctr}\right)  
\end{equation}
where $\mathbf{x}_{i}^{ctr}$ and $\hat{\mathbf{x}}_{i}^{ctr}$ are the $i$-th points corresponding to the target and predicted control points, respectively.
By regressing the $4$ control points, we can obtain the final synthesized Bézier curve to characterize the dental arch initially. 
\paragraph{GCN-based Arch Refinement}
We generate the target dental arch by connecting all the line segments that pass sequentially through the teeth centroids and then sampling uniform points from the connected line segments. The target and predicted dental arch are denoted as 
$\left\{x_i^{gt}\right\}_{i=1}^N$ and $\left\{\hat{x}_i\right\}_{i=1}^N$, respectively,
where $N$ is the number of points comprising the dental arch curve and is set to $32$. As shown in the bottom of Fig~\ref{fig:arch}, we first initialize the arch curve with uniformly-sampled points along the synthesized Bézier curve above. The nearest three votes corresponding to each initial arch point are selected, and their features are interpolated to represent the corresponding arch point features. The interpolated features are aggregated by MLP and then fed into our GCN for generating the offsets. We add the coordinates of the initial arch points and the learned offsets to generate new arch points. The learning process for generating offsets is iteratively repeated $3$ times to refine the initial dental arch prediction, generating the fine prediction of the dental arch. 
The loss function for arch points prediction can be formulated as follows:
\begin{equation}
L_{arch}=\frac{1}{N} \sum_{i=1}^{N} \ell_{1}\left(\hat{\mathbf{x}}_{i}-\mathbf{x}_{i}^{g t}\right)
\end{equation}
\subsubsection{Arch-aware Point Sampling (APS)}
With the estimated dental arch, we design an APS method to expressly select the points around the tooth crown to tackle the issue above. This is based on our observation that all teeth are sequentially arranged on a dental arch, so are their centroids. As shown in Fig.~\ref{fig:pipeline}, the APS and grouping module makes use of the predicted dental arch and generates final $N_k$ teeth proposals. Specifically, we utilize the Hungarian~\cite{kuhn1955hungarian} method to sample subsampled points in $N_m$ votes. Hungarian method considers the distances among assigned points and samples points more uniformly, compared with KNN-like methods that directly sample the K-nearest points to the dental arch. The cost matrix $ \mathbb{C} $ for Hungarian method consists of two parts: 
\begin{equation}
  \mathbb{C} = \alpha\mathbb{D_{\text{arch}}} + \beta\mathbb{D_{\text{votes}}}  
\end{equation}
The first matrix $\mathbb{D_{\text{arch}}}$ is the Euclidean distance between votes and dental arch points.
The second part $\mathbb{D_{\text{votes}}}$ is the Euclidean distance of votes displacement. $\alpha$ and $\beta$ are used to balance the importance of these two distance measurements for sampling. We experimentally set $\alpha$ and $\beta$ to be $1$ and $5$, respectively. The effect of different sampling methods on both detection and segmentation are compared, and the results are attached in the \emph{Supplementary}.

\subsubsection{Loss Function}
When training the networks, only annotations of teeth centroids are utilized. We use the Huber $\ell_{1}$ loss~\cite{ren2015faster} $L_{\text{offset}}$ to supervise the offsets prediction to obtain points $F$ from original subsampled points to their nearest annotated centroids. Next, we use Cross-Entropy loss $L_{\text{conf}}$ to supervise the proposal confidence. We assume the ground truth confidence of proposals which distances to their closest teeth centroids less than $0.3$ to be $1$ and assign the corresponding teeth centroids to the proposals such as VoteNet~\cite{qi2019deep}. In the end, base on the assigned teeth centroids, we compute the losses $L_{\text{centers}}$ and $L_{\text{boxs}}$ for learning centroids offset and regressing teeth objects box regression~\cite{qi2018frustum}. Specifically, loss for the teeth detection is as follows:
\begin{equation}
 L_{\text{det}} = L_{\text{offset}} + L_{\text{conf}} + \gamma L_{\text{centers}}   
\end{equation}
where we empirically set $\gamma$ to be $0.1$.

\subsection{Tooth Instance Segmentation}
Our segmentor is build upon PointNet++~\cite{qi2017pointnet++}. We adopt a patch-based training strategy to train the segmentor and the common cross-entropy loss function to optimize the training process. Given a centroid point, we crop the closest $M=2048$ points to the centroid point from the original point cloud $P$.
As shown in Fig.~\ref{fig:pipeline}, the input of segmentor are the backbone features and the relative coordinates to the given centroid of the cropped 3D patch, and the output is the probability mask indicating the possibility of the points from the 3D patch being tooth point. The training data for training our segmentor is all 3D patches generated by cropping the closest $M$ points to those tooth centroids of labeled tooth instances. For example, if three tooth instances are labelled in a dental model, we will generate three 3D patches by cropping the closest $M$ points to the three tooth centroids of labelled tooth instances. The patch-based training strategy can augment the training samples and fully utilize the annotation information. In the inference stage, the well-trained segmentor can segment all tooth instances of the entire dental model by fusing the segmentation results on all patches that are generated based on each detected centroid. 

\subsection{Network Training} For training the tooth centroid detection network, we sample $N = 16,000$ points uniformly from each dental model, using their 3D coordinates as the unique feature input. We first train the detection backbone in the first $210$ epochs and other network settings, such as the optimizer and the learning rate, follows ~\cite{qi2019deep}. Then we train the arch prediction branch for $100$ epochs with the fixed detection backbone. With the estimated dental arch, we perform APS to generate accurate proposals and fine-tune the network of proposal generation, as the yellow trapezoid denoted in Fig.~\ref{fig:pipeline}. Non-maximum suppression (NMS) is applied to these proposals to generate the final centroid prediction. For training the tooth instance segmentation network, tooth centroids of those annotated teeth masks in the training dental models are used to generate 3D patches by cropping the closest $M=2,048$ points to them from the corresponding point clouds. A patch-based training strategy is used for our segmentor. Our segmentator is build upon PointNet++~\cite{qi2017pointnet++} and follow the similar settings of ~\cite{qi2017pointnet++} in the training phase. All trainings are conducted under a single RTX 3090 Nvidia GPU. Please refer to \emph{Supplementary} for detailed information.

%% file: section/experiment.tex
\begin{table*}[ht]
    \centering
    \begin{tabular}{lccc|clcl}
        \toprule
        Method               & \multicolumn{3}{c|}{Tooth centroid detection} & \multicolumn{4}{c}{Tooth instance segmentation}                 \\ \hline
         & Acc. & Recall & C. Dist. & \multicolumn{2}{c}{Full} & \multicolumn{2}{c}{Weak} \\
         & & & & IoU & Dice & IoU & Dice \\ \hline
        VoteNet~\cite{qi2019deep} & $88.82$ & $85.68$  & $0.036$ & - & - & - & -  \\
        MLCVNet\cite{xie2020mlcvnet} & $90.86$ & $85.68$  & \textbf{0.033} & - & - & - & -  \\
        Group-free 3D~\cite{liu2021group} & $91.14$ & \textbf{92.70}  & $0.035$ & - & - & - & - \\\hline
        TSegNet~\cite{cui2021tsegnet} & $99.41$ & $84.94$ & $0.037$ & $94.83$ & $96.91$ & $93.39$ & $95.83$\\ 
        VoteNet \& PointNet++~\cite{qi2017pointnet++} & $84.32$ & \textbf{85.40} & $0.040$ & $93.92$ & $96.29$ & $93.38$ & $95.97$ \\
        DArch (Ours) & \textbf{99.68} & $85.39$ & $0.037$ & \textbf{95.93}  & \textbf{97.70}& \textbf{95.42} & \textbf{97.38} \\
        \bottomrule
    \end{tabular}
    \caption{Tooth centroid detection and tooth instance segmentation results compared with state-of-the-art methods in weakly- and fully annotated scenarios. "-" denotes the unavailable segmentation scores for those detection methods.}
    \label{tab:det}
\end{table*}

\if false
\begin{table*}[ht]
    \centering
    \begin{tabular}{lccc|clcl}
        \toprule
        Method               & \multicolumn{3}{c|}{Tooth centroid detection} & \multicolumn{4}{c}{Tooth instance segmentation}                 \\ \hline
        \multicolumn{1}{c}{} & \multirow{ACC} & \multirow{Recall} & \multirow{Chamfer Dist.} & \multicolumn{2}{c}{Full} & \multicolumn{2}{c}{Weak} \\ 
        \multicolumn{1}{c}{} &               &               &               & IoU & \multicolumn{1}{c}{Dice} & IoU & \multicolumn{1}{c}{Dice} \\ \hline
        VoteNet~\cite{qi2019deep} & $88.82$ & $85.68$  & $0.0360$ & - & - & - & - \\
        MLCVNet\cite{xie2020mlcvnet} & $90.86$ & $85.68$  & \textbf{0.0333} & - & - & - & -  \\
        Group-free 3D~\cite{liu2021group} & $91.14$ & \textbf{92.70}  & $0.0351$ & - & - & - & - \\
        TSegNet~\cite{cui2021tsegnet} & $99.41$ & $84.94$ & $0.0372$ & $94.83$ & $96.91$ & $93.39$ & $95.83$ \\        \hline
        VoteNet+PointNet++(random) & $87.49$ & $84.77$ & $0.0402$ & $93.09$ &  $95.69$ & $92.55$ & $95.36$  \\
        VoteNet+PointNet++(fps) & $85.59$ & $84.24$ & $0.0407$ & $91.93$ & $95.36$ & $92.51$ & $95.30$ \\
        \bottomrule
    \end{tabular}
    \caption{\textbf{Tooth centroid detection and tooth instance segmentation results compared with state-of-the-art methods in weak and full annotations cases.} "-" denotes the unavailable segmentation scores for those detection methods.}
    \label{tab:det}
\vspace{-5mm}
\end{table*}
\fi

\section{Experiments}
\subsection{Dataset and Annotation} 
\begin{figure}[t]
  \centering
   \includegraphics[width=1.0\linewidth]{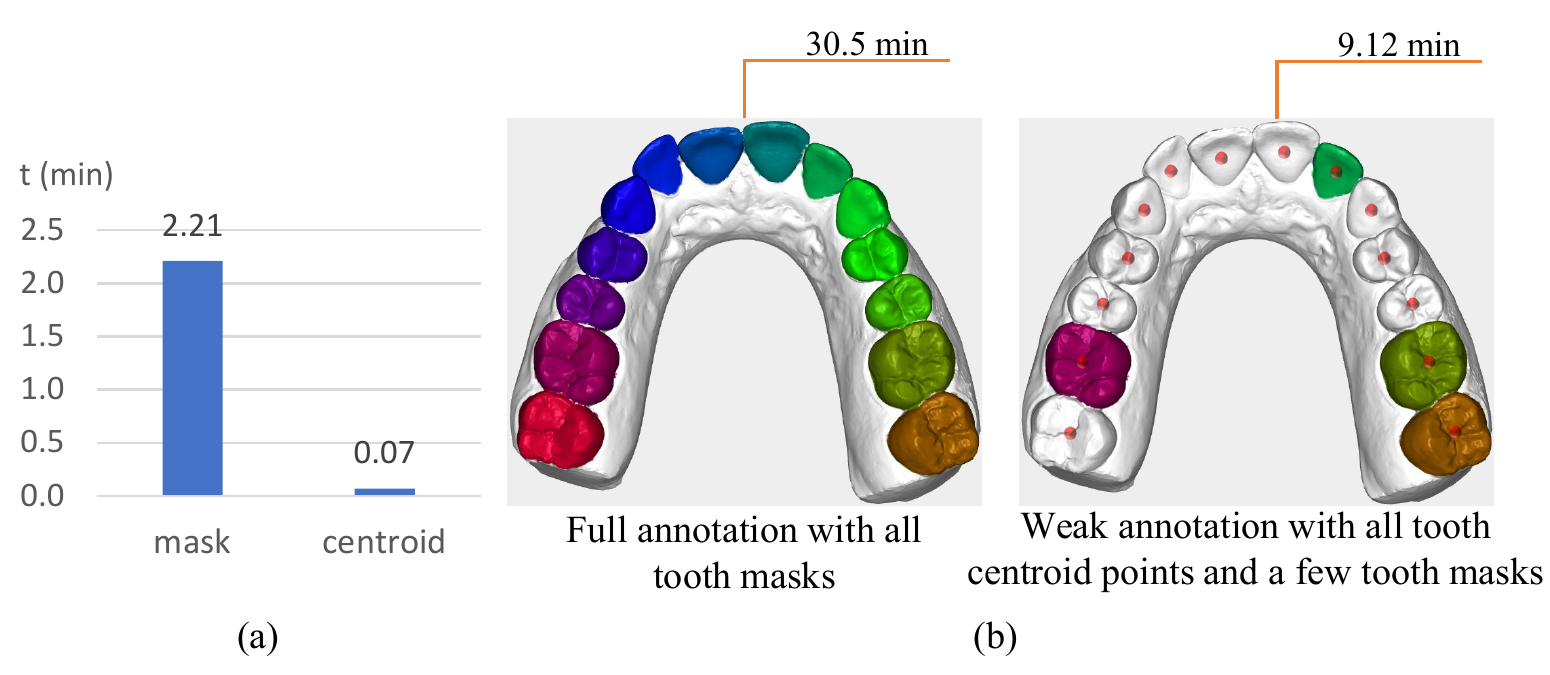}
   \caption{Illustration of time consumption for different annotation ways. (a) Comparison of time spent on labeling each tooth mask and centroid; (b) Comparison of time spent on labeling one dental model with full and weak annotations.} 
   \label{fig:time}
   \vspace{-5mm}
\end{figure}

We collected $4,773$ 3D dental models from $3,231$ patients before orthodontics. We randomly select $3,973$ models as the training models and the rest $800$ models as the test models. All training dental models contain a total of $54,658$ in teeth instances. All the dental models are fully annotated, in which all tooth instances of each dental model are manually labeled by professional dentists. In our work, we propose a low-cost annotation way, that is, labeling all tooth centroids and only a few teeth for each dental model. To calculate the time spent for full annotation and our proposed weak annotation, one of the authors manually annotates $10$ dental models with different annotation ways under the guidance of professional dentists. Although teeth centroids used in our experiments are calculated by the fully annotated teeth masks, we propose a new way to annotate teeth centroids by multi-view images, which is less time-consuming. We first render the dental model to three images of different views. Then, performed in a strict sequence, we select the center point of each tooth on these images respectively to calculate the coordinates of the to-be-annotated tooth centroid. For annotating a tooth to generate the mask, we use the popular and programmable 3D mesh editing software, Meshlab~\cite{cignoni2008meshlab}, as our annotation tool. We use the Z-painting tool provided by Meshlab by painting vertexes on each tooth instance. Fig~\ref{fig:time} shows an example of full annotation and our proposed weak annotation and indicates the averaged annotation time on one dental model for both types of annotation. As shown in Fig~\ref{fig:time}, the weak annotation way used in our work can save time to a large extent compared to the fully annotated approach used in other learning-based methods.

\subsection{Experimental Setup}
\paragraph{Competing methods.}
We compare our approach with the state-of-the-art method on tooth centroid detection and tooth instance segmentation.
As for the detection, our DArch is compared with the popular 3D detection methods(\emph{i.e.}, VoteNet~\cite{qi2019deep},  MLCVNet~\cite{xie2020mlcvnet} and Group-free 3D~\cite{liu2021group}). VoteNet is a general 3D detection method for point clouds. MLCVNet extends the VoteNet by leveraging multi-level context modules, \emph{i.e.}, patch-to-patch, object-to-object and global scene. Group-free 3D further adopts a transformer-based proposal generation networks. 
As for the segmentation, we compare our DArch with the state-of-the-art 3D tooth instance segmentation method (\emph{i.e.}, TSegNet~\cite{cui2021tsegnet}) and the combination of popular VoteNet and PointNet++. 
TSegNet is the start-of-the-art learning-based method for 3D tooth instance segmentation.\\
\textbf{Metrics.} We use the widely-used metrics-Accuracy (ACC) and Recall for evaluating the detection performance, as well as IoU and Dice metrics are used to evaluate the segmentation performance. Besides, we adopt an extra metric-Chamfer Distance~\cite{barrow1977parametric} to measure the distance between the predicted centroids and the ground truth centroids. Given two point clouds $P_{1} \subseteq R^{3}, P_{2} \subseteq R^{3}$, The Chamfer Distance can be defined as
\begin{equation}
    d_{C H}\left(P_{1}, P_{2}\right)=\sum_{x \in S_{1}} \min _{y \in P_{2}}\|x-y\|_{2}^{2}+\sum_{y \in P_{2}} \min _{x \in P_{1}}\|x-y\|_{2}^{2}
\end{equation}

\begin{figure*}[ht]
\begin{center}
  \includegraphics[width=0.99\linewidth]{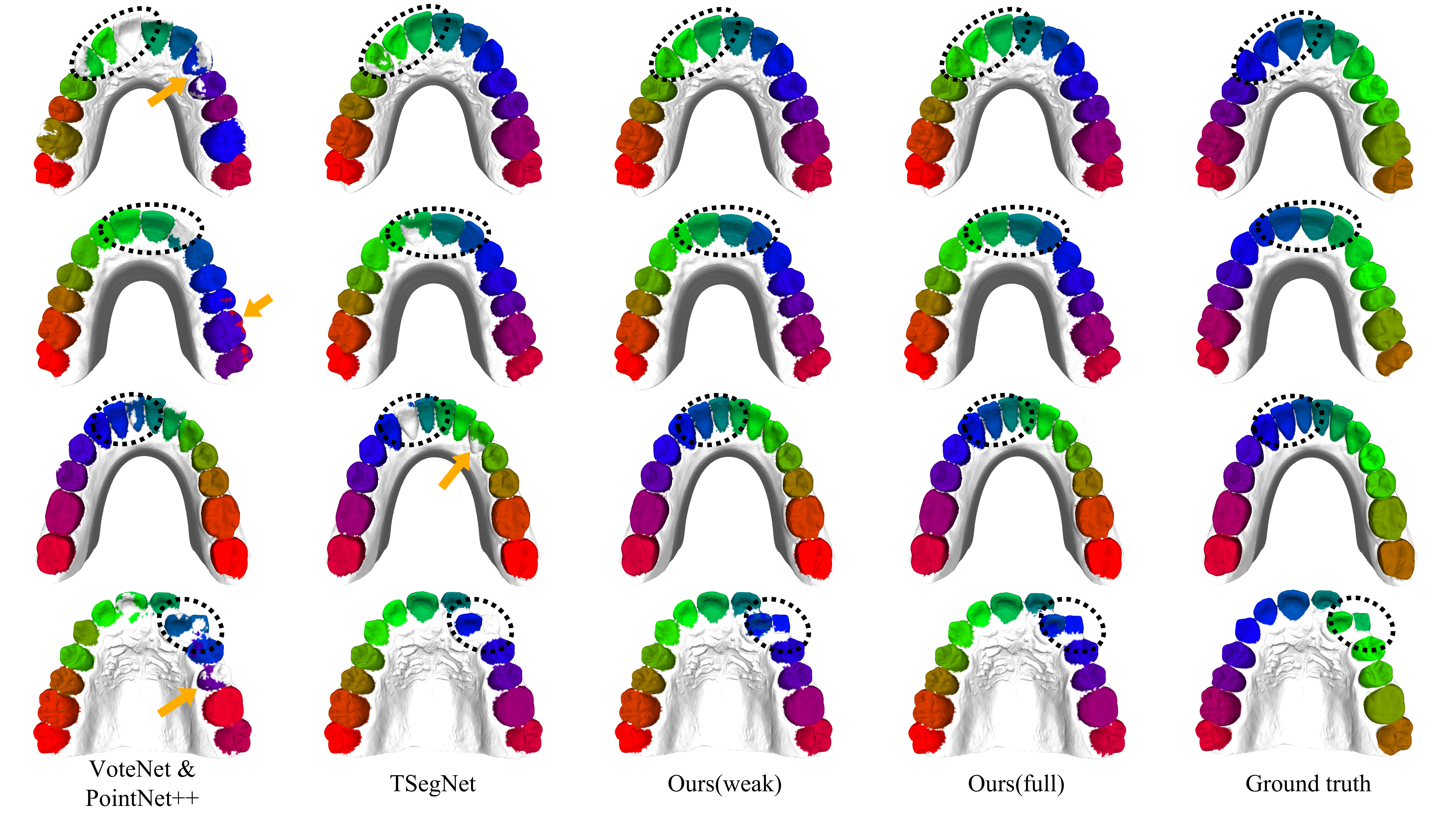}
\end{center}
  \caption{The visual comparison of dental model segmentation results produced by different methods, as well as the corresponding ground truth. From left to right are the results of other methods (1st-2nd columns) with full annotations, our results with weak annotations, our results with full annotations and the ground truth.}
\label{fig:visual}

\end{figure*}

\begin{table*}[tb]
\centering
\begin{tabular}{lcccccc}
\toprule
Method & Number & \multicolumn{3}{c}{Tooth Centroid Detection} & \multicolumn{2}{c}{Tooth Instance Segmentation} \\ \hline
           &    & Acc.   & Recall         & C. Dist.         & IoU   & Dice  \\ \cline{2-7} 
FPS        & 20 & 84.32 & 85.40           & 0.040          & 93.38 & 95.97 \\
           & 30 & 85.4  & \textbf{85.66} & 0.038          & 95.57 & 97.49 \\ \hline
APS (Ours) & 20 & 99.68 & 85.39          & \textbf{0.037} & 95.42 & 97.38 \\
       & 30     & \textbf{99.74}   & 85.37   & \textbf{0.037}  & \textbf{95.67}         & \textbf{97.53}         \\ \bottomrule

\end{tabular}%
\caption{The detection and segmentation results when using different thresholding centroid number and sampling methods. 'Number' means the number of the detected tooth centroids in the detection stage. }
\label{sample}
\end{table*}

\subsection{Comparison with Competing Methods}
\paragraph{Experimental setup.} In this section, we compare our method with different competing methods. Note that all segmentation models of our DArch and another two competing methods, TSegNet and VoteNet \& PointNet++, adopt patch-based training strategy and fuse all patch-based segmentation results to produce the segmentation result of an entire dental model. The 3D patches that are used as the input of all segmentation models are generated by cropping the closest $2,048$ points to those detected tooth centroids. As we mentioned in Section 3.2.1, the tooth centroids detected by VoteNet and our DArch are generated by NMS filtering. By thresholding, VoteNet and our DArch can generate  different numbers of the predicted tooth centroids. The number of detected tooth centroids can affect the detection and segmentation results. With a small increase in the number of the detected tooth centroids, the detection recall may increase, and the segmentation performance also may improve at the expense of decreased efficiency since the segmentation results from more patches are fused. Our experimental statistics yield an average number of the detected tooth centroids for TSegNet model of about 28.6. For fair comparison and taking into account model efficiency, we filter the proposals of VoteNet and our DArch and generate $20$ tooth centroids for both methods.\\
\textbf{Results.} The overall detection and segmentation results are presented in Table~\ref{tab:det}, and we compared these competing methods in both weakly- (\emph{i.e.}, only labeling 20\% teeth instances from all tooth instances in the training dental models) and fully annotated scenarios. Since VoteNet~\cite{qi2019deep},  MLCVNet~\cite{xie2020mlcvnet} and Group-free 3D~\cite{liu2021group} can only be used for detection, their segmentation metrics are default. From the table, we can see that our DArch achieves the best segmentation performance in both weakly- and fully-annotated scenarios. Compared to the state-of-the-art 3D tooth instance segmentation method, TSegNet, the proposed DArch improves the segmentation performance by $1.1\%$ and $0.79\%$ on the IoU and Dice, respectively, with full annotations, and by $2.03\%$ and $1.55\%$ on the IoU and Dice, respectively, with weak annotations. In the weakly-annotated scenario, our DArch improves more. The reason may be that our method can generate more accurate detection results. Owing to accurate detection results, our segmentation models perform well even in the weakly-annotated scenario. This also suggests that locating tooth objects is important for the segmentation, and our proposed weak annotation is feasible. 
The visual results of our method and other methods are shown in Fig~\ref{fig:visual}. From this figure, we can find that even only weak annotations are available for our DArch, it can also produce visually better results than other methods with full annotations, especially in areas of small teeth. 

\subsection{Ablation Studies}

\subsubsection{Sampling}
Thresholding Centroid Number and different sampling methods in the detection stage will affect the detection and segmentation performance. In this section, we investigate the effect of different thresholding centroid numbers (\emph{i.e.}, 20 and 30) and sampling methods (\emph{i.e.}, FPS and APS) on the tooth centroid detection and tooth instance segmentation. The results are reported in Table~\ref{sample}. From this table, we can observe that our proposed APS method achieves the best results in terms of the detection and segmentation results, especially the ACCs are much higher than that of other sampling methods in different centroid numbers. Besides, when the thresholding centroid number is low (\emph{i.e.}, $20$), our APS remains reflecting a relatively consistent detection and segmentation performance with the higher centroid number of $30$, while FPS decreases more. This also suggests that by leveraging dental arch prior, our APS can detect more accurate centroid points than the conventional FPS method.
\vspace{-2mm}
\if false
\begin{table}[ht]
    \centering
    \begin{tabular}{lccc|clcl}
        \toprule
        \multirow{Method} &\multicolumn{3}{c|}{Detection} &\multicolumn{4}{c}{Segmentation}\\ \hline
        &ACC & Recall & C. Dist. & IoU & Dice \\ \hline
        Random & $80.91$ & $84.63$ & $0.0410$ & $92.03$ & $95.06$  \\
        FPS & $84.32$ & \textbf{85.40} & $0.0402$ & $93.38$ & $95.97$ \\
        APS & \textbf{99.68} & $85.39$ & \textbf{0.0369} & \textbf{95.42} & \textbf{97.38} \\
        \bottomrule
    \end{tabular}
    \caption{Our tooth centroid detection and tooth instance segmentation results when using different sampling strategies with weak annotation.}
    \label{tab:sample}
\vspace{-5mm}
\end{table}
\fi

\if false
\begin{table}[]
\begin{tabular}{lllllll}
Method               &    & \multicolumn{3}{l}{Detection} & \multicolumn{2}{l}{Segmentation} \\
                     &    & Acc.     & Recall   & C. Dist.   & IoU             & Dice           \\
\multirow{2}{*}{FPS} & 20 & $84.32$   & $85.40$     & $0.0402$   & $93.38$           & $95.97$          \\
                     & 30 & $85.40$    & \textbf{85.66}    & $0.038$    & $95.57$           & $97.49$          \\
\multirow{2}{*}{APS} & 20 & $99.68$   & $85.39$    & $0.0369$   & $95.42$           & $97.38$          \\
                     & 30 & \textbf{99.74}   & $85.37$    & \textbf{0.0365}   & \textbf{95.67}           & \textbf{97.53}         
\end{tabular}
\end{table}
\fi


\subsubsection{Dental Arch Prediction}
In our work, we propose a coarse-to-fine method for predicting dental arches. We first synthesize a cubic Bézier curve using an MLP network to initially characterize the dental arch and then use a lightweight network to refine the initially estimated arch. To validate the effectiveness of the coarse-to-fine strategy, we predict the dental arch using different methods, such as direct prediction using an MLP, coarse Bézier curve regression and our proposed coarse-to-fine strategy. The results are reported in the Table~\ref{tab:ablation_line}. The results in this table indicate the effectiveness of our coarse-to-fine strategy on arch prediction. The analysis of hyperparameters is attached in the \emph{Supplementary}.
\vspace{-2mm}


\begin{table}[tb]
    \centering
    \begin{tabular}{lccc} 
    \toprule
    Method & Acc. & Recall & MSE. (1e-4) \\
    \hline
    Direct$^*$ & $93.13$ & $85.12$ & $7.50$ \\
    Coarse & $93.44$ & $85.27$ & $6.22$\\
    Coarse + Fine & \textbf{99.89} & \textbf{84.17} & \textbf{4.36} \\
    \bottomrule
    \end{tabular}
    \caption{Ablation study of Arch prediction. Direct$^*$ denotes directly predicting the arch points using an MLP network. Coarse denotes predicting the arch points only by Bézier curve regression. Fine indicates further refining the coarse prediction.}
    \label{tab:ablation_line}
    \vspace{-4mm}
\end{table}

%% file: section/conclusion.tex
\section{Conclusion}
In this work, we propose a novel tooth instance segmentation framework-\emph{DArch}. Our DArch consists of two parts of tooth centroid detection and tooth instance segmentation. This method provides a novel dental arch estimation method and introduces an arch-aware point sampling (APS) module based on the estimated dental arch for tooth centroid detection. Owing to the impressive detection performance obtained by the detection stage, our DArch has achieved superior performance to the competing segmentation methods in both weakly- and fully-annotated scenarios. Our segmentor is trained in a fully-supervised manner and does not take full advantage of the weakly-annotated centroid information and our proposed dental arch prior. In the future, we 
will design a smarter segmentor by fully leveraging this information.
\vspace{-4mm}
\paragraph{Broader Impact.} 
The segmentor of our DArch is trained in a fully-supervised manner. The training data is limited when only a small amount of teeth are manually labeled, which will limit the generalization ability of the trained segmentor. The model may generate inaccurate segmentation results on unseen dental models from the real-world, thus adversely affecting the computer-aided orthodontic treatments.
\vspace{-3mm}
\paragraph{Acknowledgements:} 
The work was supported in part by the Basic Research Project No. HZQB-KCZYZ-2021067 of Hetao Shenzhen-HK S\&T Cooperation Zone, National Key R\&D Program of China with grant No. 2018YFB1800800,  by Shenzhen Outstanding Talents Training Fund 202002, and by Guangdong Research Projects No. 2017ZT07X152 and No. 2019CX01X104. This work was also supported by NSFC-62172348, 61902334, Shenzhen General Project (JCYJ20190814112007258) and High-Performance Computing Portal under the administration of CUHK(Shenzhen) Information Technology Services Office.